\pdfoutput=1

\documentclass[11pt]{article}

\usepackage[]{acl}

\usepackage{times}
\usepackage{latexsym}

\usepackage[T1]{fontenc}

\usepackage[utf8]{inputenc}

\usepackage{microtype}

\usepackage{inconsolata}

\usepackage{times}
\usepackage{adjustbox} %
\usepackage{amsfonts}
\usepackage{amsmath}
\usepackage{amssymb}
\usepackage{array}
\usepackage{arydshln}
\usepackage{bbm}
\usepackage{booktabs}
\usepackage{comment}
\usepackage{dblfloatfix}
\usepackage{dsfont}
\usepackage{float}
\usepackage{listings}
\usepackage{graphicx,wrapfig,lipsum}
\usepackage{mathtools,nccmath}
\usepackage{multirow}
\usepackage{pifont}
\usepackage{relsize}
\usepackage{url}
\usepackage{subcaption}
\usepackage[capitalize]{cleveref}

\title{What an Elegant Bridge: Multilingual LLMs are Biased Similarly in Different Languages \\}

\author{Viktor Mihaylov \quad Aleksandar Shtedritski \\
\\
   Oxford Artificial Intelligence Society, University of Oxford
   }

\begin{document}
\maketitle
\begin{abstract}
This paper investigates biases of Large Language Models (LLMs) through the lens of grammatical gender. Drawing inspiration from seminal works in psycholinguistics, particularly the study of gender's influence on language perception, we leverage multilingual LLMs to revisit and expand upon the foundational experiments of Boroditsky (2003). Employing LLMs as a novel method for examining psycholinguistic biases related to grammatical gender, we prompt a model to describe nouns with adjectives in various languages, focusing specifically on languages with grammatical gender. In particular, we look at adjective co-occurrences across gender and languages, and train a binary classifier to predict grammatical gender given adjectives an LLM uses to describe a noun. Surprisingly, we find that a simple classifier can not only predict noun gender above chance but also exhibit cross-language transferability. We show that while LLMs may describe words differently in different languages, they are biased similarly.
\end{abstract}
 \section{Introduction} 
The way we perceive the world is not only affected by our culture \citep{oyserman2008does, masuda2008effect}, but also the language we speak \citep{boroditsky2003sex, boroditsky2001does}. 
The relationship between cognition and language has been of interest for a long time \citep{langacker1993universals}, especially through the lens of gender \citep{boroditsky2003sex, gygax2008generically}.
Recent advances in Large Language Models (LLMs), that match human performance on multiple tasks, provide an exciting opportunity to study the relationship between the psycholinguistic biases of humans and those of machines. 
While it is unclear whether the latter relationship exists, it would be a more scalable, affordable, and even ethical \cite{banyard2013ethical} alternative to human studies.

In this work, we revisit the study of \citep{boroditsky2003sex} in the era of LLMs. 
To see how grammatical gender affects cognition, \citet{boroditsky2003sex} ask speakers of languages with grammatical gender (where nouns have assigned genders) to describe various objects, finding that the language a person speaks affects the attribution of masculine or feminine characteristics to objects. 
For example, a Spanish speaker (where ``bridge'' is masculine) might describe a bridge with words like ``strong'' or "sturdy'', while a German speaker (where ``bridge'' is feminine) might use terms like ``elegant'' or ``beautiful''. 
However, several subsequent studies fail to replicate such results \citep{haertle2017does, mickan2014key, samuel2019grammatical}, which is but a symptom of the replication crisis in psychology \citep{wiggins2019replication, shrout2018psychology, maxwell2015psychology}. Similarly, studies in the field of NLP that examine the way gendered nouns are used in text corpora \citep{williams2021relationships, kann2019grammatical}, find conflicting evidence on whether there is a relationship between grammatical gender and cognition.

\begin{figure}[t]
\centering
\includegraphics[width=0.48\textwidth]{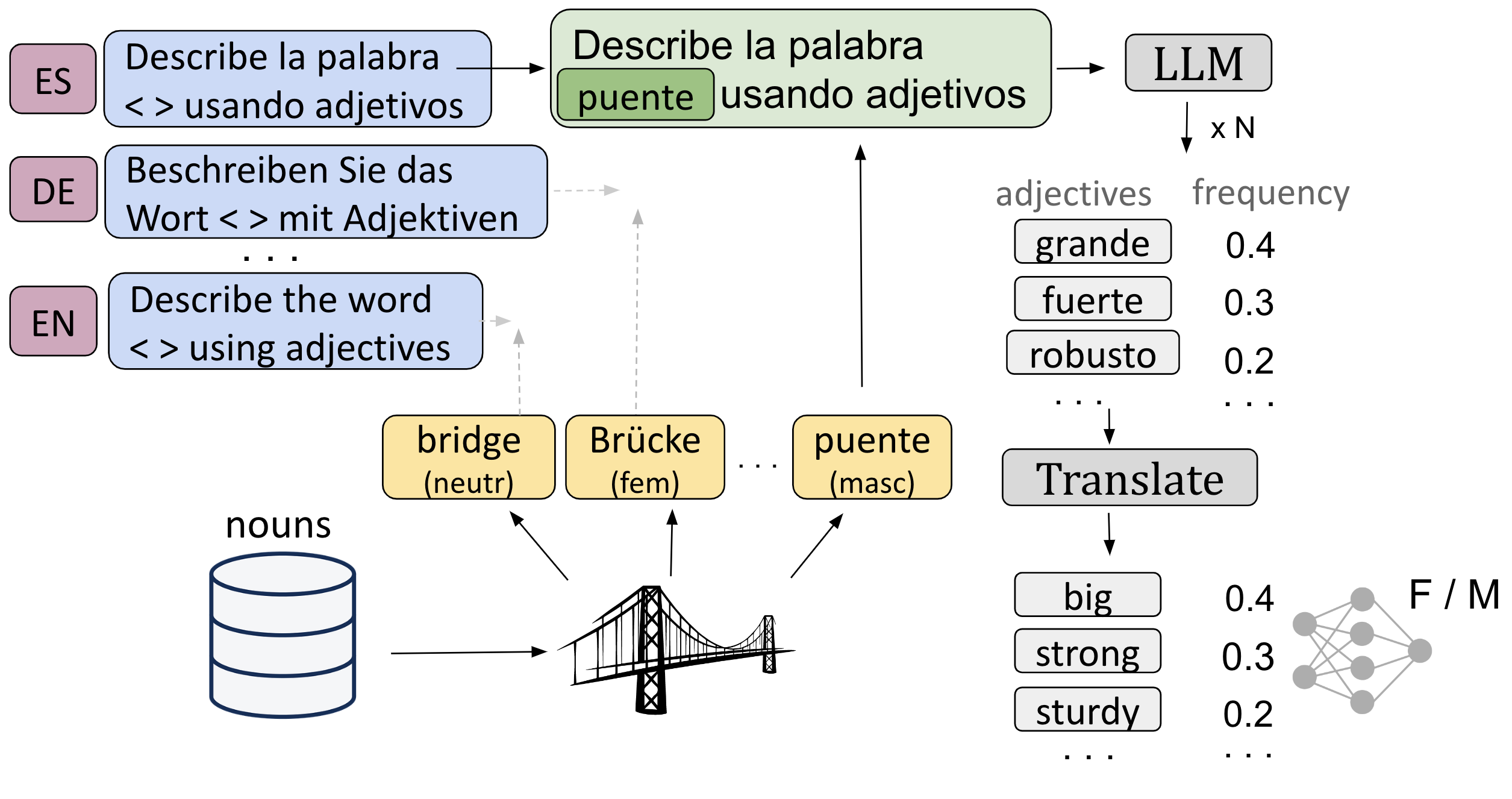}
\caption{\textbf{Probing the bias of multilingual LLMs.} We prompt a LLM to describe gendered nouns using adjectives. This allows us to study psycholinguistic biases of LLMs. For example, if the generated adjectives are predictive of the nouns's gender, we can, by training a binary classifier, predict grammatical gender by only looking at the adjectives a LLM uses to describe a word.}%
\label{fig:method}
\end{figure}

The existence of gender bias has been well studied for word embeddings \citep{bolukbasi2016man-bias-emb, basta2019evaluating-bias-emb, caliskan2017semantics-bias-emb}, as well as a range of NLP systems, such as ones for machine translation \citep{stanovsky-etal-2019-evaluating, vanmassenhove-etal-2018-getting}, image and video captioning \citep{tatman-2017-gender, hall2023visogender}, or sentiment analysis \citep{kiritchenko2018examining}. 
More recently, the social biases of LLMs have been studied \citep{kirk2021bias}.
While the multilingual capabilities of LLMs have been extensively evaluated, showing they perform well on machine translation \citep{hendy2023good-multimodal-translate, jiao2023chatgpt-multimodal-translate, wang2023document-multimodal-translate} as well as various multilingual benchmarks \citep{ahuja2023mega-multilingual-bench, bang2023multitask-multilingual-bench}, the evaluation of biases in the multilingual setting is less mature. Contrary to recent work showing that multilingual LLMs have different biases for different languages \citet{mukherjee2023global}, we find that when it comes to gendered nouns, LLMs are biased in a similar way, as the biases are predictive of each other.

In this paper, we loosely follow the protocol of \citet{boroditsky2003sex} and prompt LLMs to describe nouns using adjectives in different languages. Specifically, we focus on open-sourced LLMs (Llama-2 \citep{touvron2023llama2} and Mistral \citep{jiang2023mistral}). We select 10 languages that have grammatical gender (e.g, German and Spanish), and use the LLMs to describe gendered nouns using adjectives. This allows us to see how adjectives co-occur across languages. 
Our most important findings are that (i) a simple classifier can predict the gender of a noun using the adjectives used to describe it, and (ii) such a classifier reliably transfers across languages, suggesting LLMs are biased similarly in different languages.

 \section{Method}

In this work, we are interested in the adjectives a multilingual LLM uses to describe gendered nouns when asked in different languages. 
Here, we describe how we generate such adjectives, and how we examine whether they are predictive of the grammatical gender of the nouns.
\subsection{Describing nouns in different languages}
We show our pipeline for describing gendered nouns with adjectives in \Cref{fig:method}.
More formally, for a language $l$ we have a database of K gendered nouns $\mathcal{N}^l = \{n^l_1, n^l_2, ..., n^l_K\}$, with corresponding grammatical genders $g(n^l_i) = \{\mathrm{f}, \mathrm{m}\}$ for feminine and masculine, respectively.
We prompt the LLM to describe a noun $n^l_k$ using adjectives, which we parse into a list of $M$ adjectives $\mathcal{A}(n^l_k) = \{a^l_1, a^l_2, ..., a^l_M\}$.
For every noun $n$, we repeat the prompting $N$ times and compute the frequencies $f$ with which the adjectives appear: 
\begin{equation}
    f(a_i) = \frac{\sum_{j=1}^{N}\mathds{1} (a_i \in \mathcal{A}(n_j))}{N}.
\end{equation} 
Finally, we keep the adjectives with top-$p$ frequencies. In practice, we use $N=50$ and $p=50$.
\subsection{Predicting gender from descriptions}
\label{sec:method_predict_gender}
To examine to what extent the adjectives an LLM uses to describe a noun are predictive of its grammatical gender, we train a binary classifier $\Phi$ to predict grammatical gender: 
$$ 
\hat{g}(n_i^l) = \Phi\left(\sum_{i=1}^{p}f\left(a_i^l\right)e_g\left(a_i^l\right)\right),
$$
where the input to the classifier are  GloVe \cite{pennington-etal-2014-glove} word embeddings $e_g$ of the adjectives weighted by the adjectives frequencies $f$. In practice, we use a modified version of $f$, where $f' = - 30 / \log(f)$ to give us a better scaling. The classifier $\Phi$ is a 2-layer MLP and we train it with binary cross-entropy loss.

As shown in \Cref{fig:method}, we first translate the generated adjectives to English. We do this for two reasons. 
Firstly, adjectives in some languages are also gendered and that would help the classifier learn this shortcut (e.g. \textit{pretty} in Spanish is \textit{bonito} and \textit{bonita} for masculine and feminine, respectively). Adjectives in English are not gendered, so the classifier $\Phi$ has no way of inferring the gender of the noun from the grammatical form. Secondly, this allows for easy transfer of the classifier across languages -- e.g. we can train $\Phi$ on words generated in Hindi, and evaluate on Italian.

 \section{Experiments}

\subsection{Implementation details}

\quad \textbf{Languages} We conduct experiments on the languages Bulgarian, Czech, French, German, Greek, Hindi, Italian, Latvian, Portuguese, and Spanish.

\textbf{Nouns} We automatically collect commonly used nouns from every language, and their corresponding grammatical gender. 
For details on the way we collect those nouns, and the number of nouns per language, please refer to the Appendix.
We exclude neuter nouns as such nouns do not exist in every language.

\textbf{LLMs} In our experiments we use the open-sourced Mistral-7B \cite{jiang2023mistral} model, unless stated otherwise. We also repeat our experiments with Llama2-7B \cite{touvron2023llama2}. 

\textbf{Prompts} We prompt the LLM to describe the given noun in the corresponding language using comma-separated adjectives. In practice, we use few-shot prompts, which we show in the Appendix.

\textbf{Translation} Where we translate nouns, adjectives, or prompts, we use Google Translate \footnote{Google Translate, \url{https://translate.google.com/}}.

\subsection{Bias in generated adjectives}
\label{sec:rm}
First, we look at adjectives that commonly occur for masculine or feminine nouns.  

\begin{figure}[t]
\centering
\includegraphics[width=0.48\textwidth]{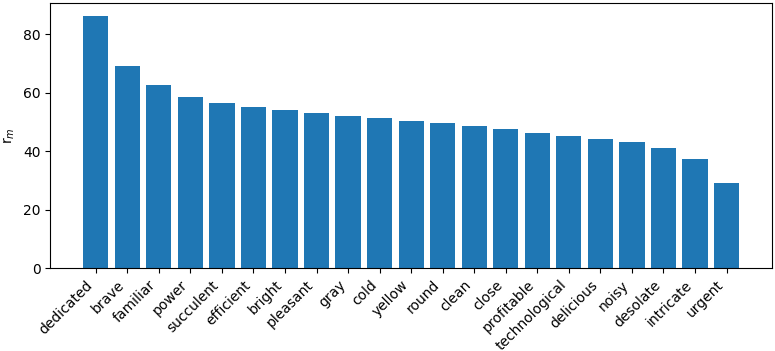}
\caption{\textbf{Bias when describing gendered nouns.} Here we prompt an LLM in Spanish and for a random sample of adjectives, show the percentage of \textit{masculine} nouns they were used for.}%
\label{fig:gendered-adjectives}
\end{figure}

For every adjective $a_i$, we look at the ratio $r_m$: 
\begin{equation}
r_m(a_i) = \frac{\sum_{n \in \mathcal{N}, g(n) = \mathrm{m}} \mathds{1} (a_i \in \mathcal{A}(n)))}{\sum_{n \in \mathcal{N}} \mathds{1} (a_i \in \mathcal{A}(n)))},
\end{equation}
which shows the proportion of masculine words it was used to describe.
We randomly sample adjectives and show their $r_m$ in~\Cref{fig:gendered-adjectives}. We see that adjectives like intricate and desolate are associated with feminine nouns, whereas adjectives like dedicated and brave are associated with masculine nouns. We show more examples for different languages in the Appendix.

\subsection{Do languages show similar biases?}
\label{sec:diff-languages-similarity}

Next, we explore whether adjectives describing masculine and feminine nouns tend to co-occur in different languages. 
To this end, we compute a gendered-adjective similarity score $S_{pq}$ for every language pair of languages $l_p$ and $l_q$. 
We do that as follows. 
We take the set of $N$ adjectives $a_1, a_2, . . ., a_N$ that  are used to describe at least 15 nouns in both $l_p$ and $l_q$. 
Then for both languages, we construct a gendered-adjective score vector $\mathbf{\sigma} \in \mathbb{R}^{N}$, where $\mathbf{\sigma}[i] = r_m(a_i)$. 
Now, $\mathbf{\sigma}_p$ and $\mathbf{\sigma}_q$ contain the gender ratio for all $N$ adjectives. 
Finally, we define the gendered-adjective similarity score $S_{pq}$ as the cosine similarity between $\mathbf{\sigma}_p$ and $\mathbf{\sigma}_q$.

\begin{figure}[t]
\centering
\includegraphics[width=0.48\textwidth]{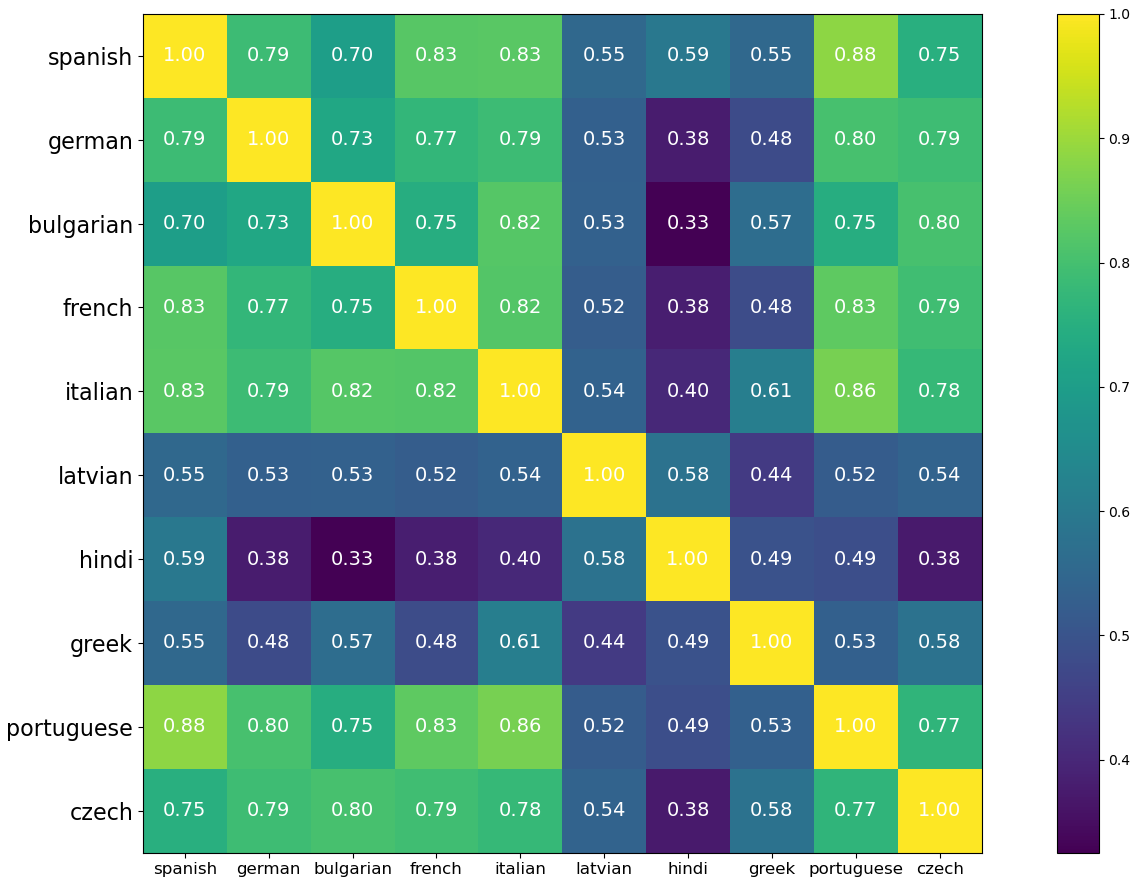}
\caption{\textbf{Gendered adjective similarity sccores}.}%
\label{fig:confusion-matrix}
\end{figure}

In \Cref{fig:confusion-matrix} we show the score $S$ for all language pairs. We see that in Romance languages (Spanish, Italian, French Portuguese), Slavic languages (Bulgarian, Czech), and Germanic languages (German), the LLM shows a high gendered-adjective similarity score, meaning that the adjectives in these languages tend to have similar value of $r_m$. On the other hand, Greek, Hindi and Latvian have a low score between themselves and others.

\subsection{Predicting the gendered nouns}
\label{sec:predict_gender}
Can we predict the gender of a noun in some language given the adjectives used to describe it? 
Following~\Cref{sec:method_predict_gender}, we train binary classifiers to predict the grammatical gender of a noun from the adjectives used to describe it (translated to English). We train a separate classifier for each language. As seen in~\Cref{tab:main-eval}, for all languages the classifier reliably does better than random -- meaning that the adjectives are predictive of gender.

\begin{table}[]
    \centering
    \footnotesize
    \begin{tabular}{lcccc}
        \toprule
        \multirow{2}{*}{Language} & \multirow{2}{*}{F1} & \multicolumn{3}{c}{Accuracy} \\
        & & Overall & Masc. & Fem.  \\
        \midrule
        Bulgarian     & 0.64 &      68.4\%      &     72.4\%    &      63.3\%      \\
        Czech         & 0.52 &      59.0\%      &     58.3\%    &      60.2\%      \\
        French        & 0.63 &      56.5\%      &     55.8\%    &      56.8\%      \\
        German        & 0.60 &      60.0\%      &     52.7\%    &      69.4\%      \\
        Greek         & 0.68 &      69.0\%      &     62.7\%    &      77.6\%      \\
        Hindi         & 0.53 &      54.3\%      &     57.5\%    &      51.2\%      \\
        Italian       & 0.46 &      68.2\%      &     73.0\%    &      54.3\%      \\
        Latvian       & 0.64 &      62.6\%      &     60.0\%    &      65.0\%      \\
        Portuguese    & 0.55 &      62.0\%      &     62.7\%    &      60.1 \%     \\
        Spanish       & 0.62 &      63.3\%      &     59.6\%    &      68.0\%      \\
        \bottomrule
        \end{tabular}
    \caption{\textbf{Predicting grammatical gender.} We train a classifier to predict the gender of nouns given the adjectives the LLM uses to describe them.}
    \label{tab:main-eval}
\end{table}

\begin{table}[t]
    \centering
    \footnotesize
    \begin{tabular}{lccccc}
        \toprule
        \multirow{2}{*}{Language} & \multirow{2}{*}{F1} & \multicolumn{4}{c}{Accuracy} \\
        & & Overall & Masc. & Fem. \\
        \midrule
        Bulgarian  & 0.56 &      62.5\%      &    64.4\%     &     59.8\%      \\
        Czech      & 0.45 &      60.6\%      &    70.6\%     &     43.5\%      \\
        French     & 0.62 &      54.8\%      &    50.3\%     &     57.3\%      \\
        German     & 0.54 &      58.6\%      &    73.1\%     &     46.0\%      \\
        Greek      & 0.64 &      60.6\%      &    47.8\%     &     75.3\%      \\
        Hindi      & 0.53 &      48.8\%      &    37.9\%     &     60.2\%      \\
        Italian    & 0.40 &      60.1\%      &    61.6\%     &     55.6\%      \\
        Latvian    & 0.41 &      51.7\%      &    81.2\%     &     29.7\%      \\
        Portuguese & 0.55 &      62.8\%      &    63.0\%     &     62.4\%      \\
        Spanish    & 0.59 &      58.8\%      &    56.7\%     &     60.1\%      \\
        \bottomrule
        \end{tabular}
    \caption{\textbf{Unseen Language Results.} We train on all other languages and predict the genders of nouns in the given language. We train a separate leave-one-out classifier for each language.}
    \label{tab:res-across-lang}
\end{table}

\begin{table}[t]
    \centering
    \resizebox{0.45\textwidth}{!}{%
    \begin{tabular}{llcccc}
        \toprule
        \multirow{2}{*}{LLM} & \multirow{2}{*}{Eval} & \multirow{2}{*}{F1} & \multicolumn{3}{c}{Accuracy} \\
        & & & Overall & Masc. & Fem. \\
        \midrule
        Mistral-7B     & Same & 0.59 &      62.3\%      &     61.5\%    &      62.6\%      \\
        Llama2-7B        & Same & 0.59 &      64.6\%      &     67.9\%    &      59.9\%      \\
        \midrule
        Mistral-7B     &Unseen& 0.53 &      57.9\%      &     60.7\%    &      55.1\%      \\
        Llama2-7B        &Unseen& 0.54 &      59.1\%      &     62.6\%    &      54.9\%      \\
        \bottomrule
        \end{tabular}
    }
    \caption{\textbf{Evaluating Llama-2.} We compare grammatical gender classifiers Llama-2 to Mistral when tested on the \textit{same} language (as in \Cref{sec:predict_gender}), or an \textit{unseen} language (as in \Cref{sec:predict-unseen}). We show mean results over all 10 languages. We see that we observe a similar predictive performance on adjectives used by Llama-2 as those by Mistral.}
    \label{tab:llama2}
\end{table}

\subsection{Transfer between languages}
\label{sec:predict-unseen}
If we train a grammatical gender classifier, like in \Cref{sec:predict_gender}, can we predict the gender of a noun in an \textbf{unseen} language? To answer this, where we train grammatical gender classifiers on adjectives from 9 languages (translated to English), and evaluate on the final language. As we see in \Cref{tab:res-across-lang}, such classifiers can reliably predict gender across languages. Interestingly, they even work better than random for Greek, Hindi and Latvian, despite the results reported in \Cref{sec:diff-languages-similarity}. We suggest that although the LLM uses different adjectives to describe masculine and feminine nouns in different languages (hence low $S_{pq}$), they are semantically similar (hence high accuracy when evaluating the classifier on an unseen language).

 \section{Discussion}
\subsection{Reproducibility}
Studying the phenomena relating cognition to grammatical gender in psychology has led to inconclusive results\citep{boroditsky2001does, haertle2017does, mickan2014key, samuel2019grammatical}. These could be explained by different experimental settings with speakers of different languages, which are difficult to control in a human study. Similarly, prior works that examine text corpora using NLP techniques show conflicting results \citep{williams2021relationships, kann2019grammatical}. The results of these works heavily depend on the text corpora analyzed, and the methods used to identify adjective-noun pairs, which might be subpar for languages other than English. Our method presents more consistent results by ensuring consistent evaluation across languages.

\subsection{Importance of our results}
Our results are only valid for noun-adjective associations in LLMs. However, these associations have been learnt through co-occurences of these words in text corpora, which have been produced by speakers of the respective languages. Future work should study how well such biases in LLMs are predictive of biases of humans.

The results we present suggest a consistent bias that associates nouns with adjectives, depending on their grammatical gender. This could be important when LLMs are used to describe humans using objects, or vice versa (anthropomorphism, personification, metaphors, ...), where traits of these objects are transferred to the human. Furthermore, using LLMs to perform machine translation of such phrases could lead to a loss of meaning or unexpected biases.

\section{Conclusion}
In this work, we revisit the psycholinguistic experiments of \citet{boroditsky2003sex}, confirming the hypothesis of their work applies to LLMs, where different words are used to described masculine and feminine nouns. Our most surprising finding is that we can reliably zero-shot transfer a classifier that predicts grammatical gender across languages. This shows that while LLMs might think differently on different languages, they are biased similarly when it comes to grammatical gender. We hope this work inspires others to explore psycholonguistic experiments applied to LLMs, and to drive a discussion of whether such results can be useful to inform or motivate human experiments. 

\section{Limitations}
We only conducted experiments and observed these effects for the opens-sourced Mistral-7B and Llama2-7B models. It is not clear if similar effects can be observed in larger LLMs, or commercial LLMs such as GPT-4. While we ensured to cover a wide range of languages, the ones we used are by no means exhaustive and only cover indo-european languages. Finally, we only explore the biases of general-purpose, multilingual LLMs. Looking into specialised LLMs, fine-tuned for the specific language, might be more representative of what models would be used in practice. 

\section{Acknowledgements} This work has been completed with the support of the Oxford AI student society, the EPSRC Centre for Doctoral Training in Autonomous Intelligent Machines \& Systems [EP/S024050/1] (A.S.) and compute resources from the AWS Responsible AI Grant.

\bibliography{anthology,refs}
\clearpage
\appendix

\label{sec:appendix}

\begin{center}
	\textbf{\Large Appendix}
\end{center}
\section{Collecting nouns}
We collect words in German \footnote{\url{https://frequencylists.blogspot.com/2016/01/the-2980-most-frequently-used-german.html}} and Spanish \footnote{\url{https://frequencylists.blogspot.com/2015/12/the-2000-most-frequently-used-spanish.html}} from a blog post that lists commonly used words in these languages, and shows their grammatical gender. For 
Bulgarian \footnote{\url{https://en.wiktionary.org/wiki/Category:Bulgarian_nouns_by_gender}}, 
Greek \footnote{ \url{https://en.wiktionary.org/wiki/Category:Greek_nouns_by_gender}},
Czech \footnote{\url{https://en.wiktionary.org/wiki/Category:Czech_nouns_by_gender}},
French \footnote{\url{https://en.wiktionary.org/wiki/Category:French_nouns_by_gender}},
Hindi \footnote{\url{https://en.wiktionary.org/wiki/Category:Hindi_nouns_by_gender}},
Italian \footnote{\url{https://en.wiktionary.org/wiki/Category:Italian_nouns_by_gender}},
Latvian \footnote{\url{https://en.wiktionary.org/wiki/Category:Latvian_nouns_by_gender}}
and Portuguese \footnote{\url{https://en.wiktionary.org/wiki/Category:Portuguese_nouns_by_gender}}, we take a list of words and their grammatical gender from Wikipedia. Following that, we only select words whose English translation is in the list of commonly used words in either German or Spanish. 

\begin{table}[h]
    \centering
    \footnotesize
    \begin{tabular}{lcccc}
        \toprule
        Language   & Total & Masc. & Fem. \\
        \midrule
        Bulgarian  & 1414  & 839  &  575   \\
        Czech      & 2383  & 1501 &  882   \\
        French     & 2763  & 996  &  1767  \\
        German     & 2031  & 952  &  1089  \\
        Greek      & 1257  & 670  &  587   \\
        Hindi      & 830   & 425  &  405   \\
        Italian    & 2919  & 2219 &  700   \\
        Latvian    & 1223  & 522  &  701   \\
        Portuguese & 1766  & 1119 &  647   \\
        Spanish    & 1758  & 896  &  862   \\
        \bottomrule
        \end{tabular}
    \caption{\textbf{Dataset Statistics.} We present the number of masculine and feminine words we consider for all 10 languages. The languages are sorted alphabetically.}
    \label{tab:dataset-stats}
\end{table}

We show the number of collected nouns per language in~\Cref{tab:dataset-stats}. 
We use 90\% of the nouns in each language for training, and 10\% for testing.

\begin{table}[]
    \centering
    \resizebox{0.45\textwidth}{!}{%
    \begin{tabular}{lcccc}
        \toprule
        \multirow{2}{*}{LLM}& \multirow{2}{*}{F1} & \multicolumn{3}{c}{Accuracy} \\
        & & Overall & Male & Female  \\
        \midrule
        Mistral-7B  &  0.57 &      55.0\%      &     50.0\%    &      60.0\%      \\
        Llama2-7B    &  0.70 &      65.0\%      &     50.0\%    &      80.0\%      \\
        \bottomrule
        \end{tabular}
    }
    \caption{\textbf{Evaluating the agreement with native English.} We evaluate the agreement of our classifier trained on 10 gendered languages to the perceived grammatical gender of native English speakers, which we treat as ground truth.}
    \label{tab:human-exp}
\end{table}

\section{Excluding animate nouns}
Following prior works that look into grammatical gender by looking at word co-occurrence in text corpora \citep{williams2021relationships}, we exclude animate nouns from our datasets in all languages (e.g. ``uncle'', ``cashier'', ``engineer'', etc.). We repeat the experiments from \Cref{sec:predict_gender} in \Cref{tab:main-eval-without-gendered}, and see that the inclusion of animate nouns does not affect overall results. 

\begin{table}[h]
    \centering
    \footnotesize
    \begin{tabular}{lcccc}
        \toprule
        \multirow{2}{*}{Language} & \multirow{2}{*}{F1} & \multicolumn{3}{c}{Accuracy} \\
        & & Overall & Masc. & Fem.  \\
        \midrule
        Bulgarian     & 0.70 &      71.1\%      &     73.8\%    &      68.3\%      \\
        German        & 0.69 &      63.8\%      &     63.1\%    &      64.2\%      \\
        Spanish       & 0.56 &      55.3\%      &     56.2\%    &      54.4\%      \\
        Italian       & 0.51 &      65.2\%      &     64.5\%    &      67.1\%      \\
        Czech         & 0.55 &      57.2\%      &     54.3\%    &      61.2\%      \\
        Greek         & 0.68 &      69.5\%      &     79.6\%    &      60.1\%      \\
        Portuguese    & 0.60 &      61.1\%      &     56.7\%    &      67.2\%      \\
        Hindi         & 0.59 &      58.1\%      &     67.7\%    &      51.2\%      \\
        Latvian       & 0.70 &      63.2\%      &     60.0\%    &      64.8\%      \\
        French        & 0.60 &      57.0\%      &     58.8\%    &      55.8\%      \\
        \bottomrule
        \end{tabular}
    \caption{\textbf{Gendered Nouns Predictions.} This table is for the filtered dictionaries, i.e. without jobs/mother/father etc.}
    \label{tab:main-eval-without-gendered}
\end{table}

\begin{figure}[h!]
\includegraphics[width=0.45\textwidth]{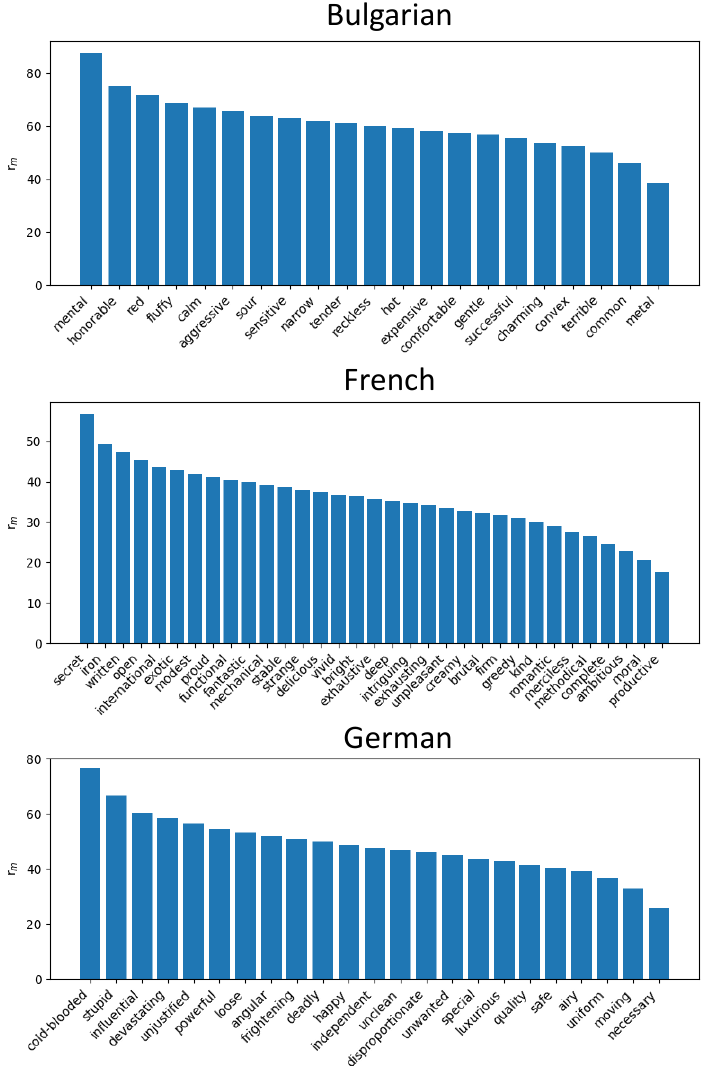}
\caption{\textbf{Bias when describing gendered nouns}. Here we prompt an LLM in Bulgarian, French, and German and for a random sample of adjectives, show the percentage of masculine nouns they were used for.}%
\label{fig:rm-app}
\end{figure}

\section{Gendered adjectives}
We show more examples of adjectives that are predominantly used for masculine (or feminine) nouns in \Cref{fig:rm-app}, similarly to \Cref{sec:rm}.

\section{Promps}

The prompt we use in English is as follows:

\small{\texttt{
***Question***: Describe the word ``bottle'' using comma-separated adjectives.
***Answer***: glass, sleek, thin, brittle, elegant, transparent, clear, tall, fragile, shiny\newline
***Question***: Describe the word ``stone'' using comma-separated adjectives.
***Answer***: round, old, strong, cold, solid, ancient, sturdy, dense, natural, durable\newline
***Question***: Describe the word <> using comma-separated adjectives.
***Answer***:}}

\normalsize{For the other languages we translate the prompt, e.g. in Spanish we use:} 

\small{\texttt{***Pregunta***: Describe la palabra ``botella'' usando adjetivos separados por comas.
***Respuesta***: vidrio, liso, delgado, quebradizo, elegante, transparente, claro, alto, frágil, brillante\newline
***Pregunta***: Describe la palabra ``piedra'' usando adjetivos separados por comas.
***Respuesta***: redondo, viejo, fuerte, frío, sólido, antiguo, robusto, denso, natural, duradero\newline
***Pregunta***: Describe la palabra <>  usando adjetivos separados por comas.
***Respuesta***:}}

\end{document}